\pdfoutput=1
\documentclass[11pt]{article}

\usepackage[final]{acl} 

\usepackage{times}
\usepackage{latexsym}

\usepackage{comment}
\usepackage{amssymb}
\usepackage{amsmath}
\usepackage{pifont}
\usepackage{color}
\usepackage{xcolor}

\usepackage{CJKutf8}

\usepackage{multirow}
\usepackage{booktabs}
\usepackage{hyperref}
\usepackage{breakurl}

\usepackage{booktabs}   %add
\usepackage{graphicx}
\usepackage{float}
\usepackage{comment}
\usepackage{natbib}

\usepackage[subrefformat=parens]{subcaption}
\usepackage{enumitem}

\usepackage{array}

\usepackage[T1]{fontenc}
\usepackage[utf8]{inputenc}

\usepackage{microtype}

\usepackage{inconsolata}

\title{Comprehensive Evaluation of Large Language Models for Topic Modeling}

\author{
  Tomoki Doi \\
  The University of Tokyo \\
  \texttt{\normalsize{doi-tomoki701@is.s.u-tokyo.ac.jp}} \\\And
  Masaru Isonuma\\
  The University of Edinburgh\\
  \texttt{\normalsize{m.isonuma@ed.ac.uk}} \\\And
  Hitomi Yanaka \\
  The University of Tokyo \\
  \texttt{\normalsize{hyanaka@is.s.u-tokyo.ac.jp}}\\
}

\begin{document}
\maketitle
\begin{abstract}
Recent work utilizes Large Language Models (LLMs) for topic modeling, generating comprehensible topic labels for given documents.
However, their performance has mainly been evaluated qualitatively, and there remains room for quantitative investigation of their capabilities.
In this paper, we quantitatively evaluate LLMs from multiple perspectives: the quality of topics, the impact of LLM-specific concerns, such as hallucination and shortcuts for limited documents, and LLMs' controllability of topic categories via prompts.
Our findings show that LLMs can identify coherent and diverse topics with few hallucinations but may take shortcuts by focusing only on parts of documents.
We also found that their controllability is limited.

\end{abstract}

\section{Introduction}
Recent studies have introduced topic modeling using prompt-based LLMs~\cite{openai_instructgpt_2022, openai_gpt4_2023}, which directly generate comprehensible topic descriptions in given documents.
Specifically, \citet{pham_topicgpt_2023} shows that topics generated by LLMs are highly consistent with human-annotated topic labels, and \citet{mu_lrec_2024} demonstrates that well-crafted prompts could control the granularity of the generated topics.

However, since previous studies primarily evaluate the quality of  topics through human judgment, it remains unclear to what extent the topics generated by LLMs meet the criterion of coherence and diversity considered in the conventional topic modeling.
In addition, LLM-specific concerns have not been considered enough in the previous evaluations; it is not impractical to manually examine the impact of hallucinations and shortcuts, in which LLMs reflect only parts of documents.
%Neglecting these aspects could result in inaccurate interpretations of topics in provided documents.
%
LLMs may generate topics that induce misinterpretation of documents due to these issues, and they may not necessarily be superior to conventional topic models.

In this paper, we quantitatively evaluate LLMs on topic modeling from multiple perspectives: the quality of generated topics, the risk of hallucinations and shortcuts, and their controllability.
We conduct experiments employing two prompting methods on three datasets from various domains.
To assess the quality of topics generated by LLMs, we directly compare LLMs with conventional topic models using well-established metrics: topic coherence and diversity.
Additionally, we introduce \textit{document coverage} and \textit{factuality} to assess the impact of hallucinations and shortcuts; document coverage indicates how many documents are covered by the generated topics, and factuality evaluates the extent to which the topics exist in the documents.
Furthermore, we explore the controllability of LLMs by examining how effectively the category of the generated topics can be controlled through prompts.

The contributions of this study are as follows:
\vspace{-0.5\baselineskip}
\begin{enumerate}
    \setlength{\parskip}{0.15\baselineskip}
    \setlength{\itemsep}{0.15\baselineskip}
    \item We validate the performance of LLMs for topic modeling by comparing them with conventional models and show that LLMs can identify more coherent topics while maintaining the diversity of topics.
    \item We assess the document coverage and factuality of topics generated from LLMs. Experiments indicate few issues with hallucination but suggest the possibility of taking shortcuts by focusing only on parts of given documents.
    \item We examine the controllability of LLMs in generating topics associated with a predefined category specified in prompts, quantitatively demonstrating their limitations.%providing a quantitative assessment of the current state of this capability.
\end{enumerate}

\section{Related Work}
\label{sec:background}
Topic modeling is a classical task of discovering latent topics that best describe a set of documents~\cite{blei_lda_2003,churchill_topicmodel_evolution_2022}.
While traditional latent Dirichlet allocation (LDA, \citealp{blei_lda_2003}) was the primary approach, neural topic models and clustering-based methods using contextual embeddings have shown impressive performance \cite{srivastava_prodlda_2017,dieng_etm_2020,zhang_neural_clustering_2022}.

Recent work \cite{pham_topicgpt_2023,mu_lrec_2024} has introduced prompt-based approaches using LLMs for topic modeling, which generates labels or brief sentences as topics, instead of conventional topic words (i.e., a set of words in documents).
To leverage input length constraints of LLMs, 
\citet{mu_lrec_2024} uses a parallel strategy, where LLMs identify topics for each document and then merge them.
\citet{pham_topicgpt_2023} employs a sequential strategy that supplies LLMs with each document along with previously identified topics and iteratively updates the topic list.
These studies primarily evaluated topics generated by LLMs through human judgment based on topic visualization \cite{mu_lrec_2024} and ground-truth labels \cite{pham_topicgpt_2023}.
It has been indicated that they could better align with human preferences compared to conventional topic models. %and could be guided through instructions in prompts.
However, previous studies do not consider quantitative assessment of the coherence and diversity of topics.
In addition, they do not investigate LLM issues, such as hallucination and shortcuts.
%the extent of these capabilities has not been clarified, and qualitative evaluation may not adequately address LLM issues, such as hallucination and shortcuts.
%that reflect only specific documents, as it requires referencing extensive documents.

% TODO

\section{Topic Modeling with LLMs}
\label{Topic Modeling by GPT-4}
Inspired  by the previous work \cite{pham_topicgpt_2023, mu_lrec_2024}, we consider two approaches to perform topic modeling with LLMs: parallel prompting and sequential prompting.
We use the three prompts shown in Table 1.
For these approaches, we apply common preprocessing, which involves randomly splitting a document set into subsets of the same size, smaller than the context length of the LLMs.
%applied to our methods.
To facilitate detailed evaluations, we adopt topic word representation for generated topics.

\paragraph{Parallel Prompting}
In the parallel prompting, LLMs with \textbf{Par$_\text{TM}$} prompt execute topic modeling for each subset in parallel.
Then, the topics of each subset are merged by LLMs with \textbf{Par$_\text{Mrg}$} prompt.
%We use two kinds of prompts shown in Table \ref{table:prompts}: (i) \textbf{Par$_\text{TM}$} prompt for parallel topic modeling for each subset, and (ii) \textbf{Par$_\text{Mrg}$} prompt for merging topics of each results.

\paragraph{Sequential Prompting}
In the sequential prompting, LLMs identify topics for each subset sequentially, considering the previously identified topics. 
We use \textbf{Par$_\text{TM}$} prompt for the first subset, then use \textbf{Seq$_\text{TM}$} prompt for the other subsets. 
%This prompt contains topics identified in the prior subset and instructions to refer to them.

\begin{table}[t!]
    \setlength{\tabcolsep}{3pt}
    \small
    \center
    \scalebox{0.95}[0.95]{
    \begin{tabular}{cl}
    \toprule
     ID & Prompt \\
    \midrule
    \textbf{Par$_\text{TM}$} &  Write the results of simulating topic modeling\\
         &  for the following documents: \texttt{[DOCS]}. \\
         
    \rule{0pt}{2.5ex}  
    
    \textbf{Par$_\text{Mrg}$} &  Write the results of merging the following \\
        &  topic modeling results:\texttt{[TOPICS]},\texttt{[TOPICS]}, ...\\
        
    \rule{0pt}{2.5ex}  
    
    \textbf{Seq$_\text{TM}$} &  Write the results of simulating topic modeling\\
        &  for the following documents: \texttt{[DOCS]}, Make the\\
        &  most use of the following topics: \texttt{[TOPICS]}. \\
    
    \rule{0pt}{2.5ex}  
    
    \textbf{Base/Orcl} &  Discover latent \texttt{[K]} topics (specifically related to\\
    (\textbf{Ctrl$_\texttt{[CAT]}$})& \texttt{[CAT]}) in the following documents: \texttt{[DOCS]}.\\
        
    \bottomrule
    \end{tabular}
    }
    \caption{Prompts used in this study. \texttt{[DOCS]} and \texttt{[TOPICS]} are replaced by a subset of documents and by previously identified topics, respectively. \texttt{[CAT]} is used to specify the target category.
    %See Appendix \ref{sec:appendix_examples_of_prompts} for concrete prompt examples.
    }
    \label{table:prompts}
\end{table}
%such as ``\texttt{\# handware startup ... making\textbackslash n \# bomb ...}'' 
%such as ``\texttt{Topic 1: irs rule exempt ...\textbackslash n Topic 2: ...}

%\input{components/table_datasets_statistics}
% \usepackage{multirow}
% \usepackage{graphicx}
%\sisetup{detect-weight=true,detect-inline-weight=math}
\begin{table*}[t!]
\small
\resizebox{\textwidth}{!}{%
%\begin{tabular}{lSSSSrSSSSrSSSS}
\begin{tabular}{lllllrllllrllll}
\hline
\multicolumn{1}{c}{\multirow{3}{*}{Model}} &
  \multicolumn{4}{c}{Tweet} &
  \multicolumn{1}{c}{} &
  \multicolumn{4}{c}{GoogleNewsT} &
  \multicolumn{1}{c}{} &
  \multicolumn{4}{c}{StackOverFlow} \\ \cline{2-5} \cline{7-10} \cline{12-15} 
\multicolumn{1}{c}{} &
  \multicolumn{2}{c}{$K=5$} &
  \multicolumn{2}{c}{$K=15$} &
  \multicolumn{1}{c}{} &
  \multicolumn{2}{c}{$K=5$} &
  \multicolumn{2}{c}{$K=15$} &
  \multicolumn{1}{c}{} &
  \multicolumn{2}{c}{$K=5$} &
  \multicolumn{2}{c}{$K=15$} \\
\multicolumn{1}{c}{} &
  \multicolumn{1}{c}{\textit{Cv}} &
  \multicolumn{1}{c}{\textit{TU}} &
  \multicolumn{1}{c}{\textit{Cv}} &
  \multicolumn{1}{c}{\textit{TU}} &
  \multicolumn{1}{c}{\textit{\textbf{}}} &
  \multicolumn{1}{c}{\textit{Cv}} &
  \multicolumn{1}{c}{\textit{TU}} &
  \multicolumn{1}{c}{\textit{Cv}} &
  \multicolumn{1}{c}{\textit{TU}} &
  \multicolumn{1}{c}{\textit{\textbf{}}} &
  \multicolumn{1}{c}{\textit{Cv}} &
  \multicolumn{1}{c}{\textit{TU}} &
  \multicolumn{1}{c}{\textit{Cv}} &
  \multicolumn{1}{c}{\textit{TU}} \\ \hline
LDA &
  0.394 &
  0.800 &
  0.401 &
  0.568 &
   &
  0.426 &
  0.984 &
  0.406 &
  0.963 &
   &
  0.320 &
  0.928 &
  0.425 &
  0.883 \\
LDA$_{\rm{Aug}}$ &
  0.445 &
  0.968 &
  0.436 &
  0.856 &
  \textbf{} &
  0.411 &
  0.984 &
  0.381 &
  0.981 &
  \textbf{} &
  0.360 &
  0.920 &
  0.508 &
  0.952 \\
TSCTM &
  0.393 &
  \bfseries 1.000 &
  0.467 &
  0.997 &
  \multicolumn{1}{l}{} &
  0.333 &
  \bfseries 1.000 &
  0.374 &
  \bfseries 1.000 &
  \multicolumn{1}{l}{} &
  0.244 &
  \bfseries 1.000 &
  0.313 &
  \bfseries 1.000 \\
TSCTM$_{\rm{Aug}}$ &
  0.355 &
  \bfseries 1.000 &
  0.433 &
  \bfseries 1.000 &
  \multicolumn{1}{l}{} &
  0.243 &
  \bfseries 1.000 &
  0.346 &
  \bfseries 1.000 &
  \multicolumn{1}{l}{} &
  0.218 &
  \bfseries 1.000 &
  0.276 &
  \bfseries 1.000 \\
BERTopic &
  0.514 &
  \bfseries 1.000 &
  0.537 &
  \bfseries 1.000 &
   &
  0.439 &
  \bfseries 1.000 &
  0.437 &
  \bfseries 1.000 &
   &
  0.459 &
  \bfseries 1.000 &
  0.485 &
  0.971 \\
BERTopic$_{\rm{Aug}}$ &
  0.535 &
  \bfseries 1.000 &
  0.526 &
  \bfseries 1.000 &
   &
  0.412 &
  \bfseries 1.000 &
  0.417 &
  \bfseries 1.000 &
   &
  0.460 &
  \bfseries 1.000 &
  0.489 &
  0.955 \\ \hline\hline
\bfseries GPT-3.5$_{\rm{Par}}$ &
  0.476 &
  0.992 &
  0.532 &
  0.900 &
   &
  0.571 &
  0.960 &
  0.535 &
  0.913 &
   &
  0.312 &
  0.864 &
  0.496 &
  0.913 \\
\bfseries GPT-3.5$_{\rm{Seq}}$ &
  0.552 &
  0.960 &
  0.515 &
  0.920 &
   &
  0.562 &
  0.984 &
  0.489 &
  0.948 &
   &
  0.441 &
  0.896 &
  0.517 &
  0.775 \\ \hline
\bfseries GPT-4$_{\rm{Par}}$ &
  0.562 &
  \bfseries 1.000 &
  \bfseries 0.576 &
  0.971 &
  \multicolumn{1}{l}{} &
  \bfseries 0.618 &
  0.976 &
  0.532 &
  0.925 &
  \multicolumn{1}{l}{} &
  \bfseries 0.466 &
  0.904 &
  \bfseries 0.571 &
  0.864 \\ 
\bfseries GPT-4$_{\rm{Seq}}$ &
  \bfseries 0.577 &
  0.992 &
  0.551 &
  0.976 &
  \multicolumn{1}{l}{} &
  0.556 &
  0.944 &
  \bfseries 0.561 &
  0.963 &
  \multicolumn{1}{l}{} &
  0.318 &
  0.744 &
  0.532 &
  0.853 \\
  \hline
\end{tabular}%
}
\caption{Topic coherence (\textit{Cv}) and diversity (\textit{TU}) results under 5 and 15 topics ($K=5$ and $K=15$).  LLM$_\text{Par}$ and LLM$_\text{Seq}$ mean LLMs with the parallel and sequential prompting, respectively. MODEL$_{\rm{Aug}}$ means the performance of the model with data augmentation. The maximum \textit{TU} is 1.000 when topic words are totally distinct from each other. The best scores are in \textbf{bold}.}
\label{tab:Coherence-and-Diversity-for-ACL2024}
\end{table*}

% TO DO
% \usepackage{multirow}
% \usepackage{graphicx}
\begin{table*}[t!]
\small
\resizebox{\textwidth}{!}{%
\begin{tabular}{lllllrllllrllll}
\hline
\multicolumn{1}{c}{\multirow{3}{*}{Model}} &
  \multicolumn{4}{c}{Tweet} &
  \multicolumn{1}{c}{} &
  \multicolumn{4}{c}{GoogleNewsT} &
  \multicolumn{1}{c}{} &
  \multicolumn{4}{c}{StackOverFlow} \\ \cline{2-5} \cline{7-10} \cline{12-15} 
\multicolumn{1}{c}{} &
  \multicolumn{2}{c}{$K=5$} &
  \multicolumn{2}{c}{$K=15$} &
  \multicolumn{1}{c}{} &
  \multicolumn{2}{c}{$K=5$} &
  \multicolumn{2}{c}{$K=15$} &
  \multicolumn{1}{c}{} &
  \multicolumn{2}{c}{$K=5$} &
  \multicolumn{2}{c}{$K=15$} \\
\multicolumn{1}{c}{} &
  \multicolumn{1}{c}{\textit{DC}} &
  \multicolumn{1}{c}{\textit{Fa}} &
  \multicolumn{1}{c}{\textit{DC}} &
  \multicolumn{1}{c}{\textit{Fa}} &
  \multicolumn{1}{c}{\textit{\textbf{}}} &
  \multicolumn{1}{c}{\textit{DC}} &
  \multicolumn{1}{c}{\textit{Fa}} &
  \multicolumn{1}{c}{\textit{DC}} &
  \multicolumn{1}{c}{\textit{Fa}} &
  \multicolumn{1}{c}{\textit{\textbf{}}} &
  \multicolumn{1}{c}{\textit{DC}} &
  \multicolumn{1}{c}{\textit{Fa}} &
  \multicolumn{1}{c}{\textit{DC}} &
  \multicolumn{1}{c}{\textit{Fa}} \\ \hline
LDA &
  \bfseries 0.337 &
  \bfseries 1.000 &
  0.561 &
  \bfseries 1.000 &
   &
  0.488 &
  \bfseries 1.000 &
  0.664 &
  \bfseries 1.000 &
   &
  \bfseries 0.684 &
  \bfseries 1.000 &
  \bfseries 0.842 &
  \bfseries 1.000 \\
LDA$_{\rm{Aug}}$ &
  0.307 &
  \bfseries 1.000 &
  \bfseries 0.579 &
  0.997 &
  \textbf{} &
  \bfseries 0.531 &
  \bfseries 1.000 &
  \bfseries 0.763 &
  \bfseries 1.000 &
  \textbf{} &
  0.659 &
  \bfseries 1.000 &
  0.838 &
  \bfseries 1.000 \\
TSCTM &
  0.176 &
  \bfseries 1.000 &
  0.388 &
  \bfseries 1.000 &
  \multicolumn{1}{l}{} &
  0.405 &
  \bfseries 1.000 &
  0.740 &
  \bfseries 1.000 &
  \multicolumn{1}{l}{} &
  0.141 &
  \bfseries 1.000 &
  0.480 &
  \bfseries 1.000 \\
TSCTM$_{\rm{Aug}}$ &
  0.187 &
  \bfseries 1.000 &
  0.331 &
  0.987 &
  \multicolumn{1}{l}{} &
  0.309 &
  \bfseries 1.000 &
  0.608 &
  0.979 &
  \multicolumn{1}{l}{} &
  0.419 &
  0.888 &
  0.441 &
  0.888 \\
BERTopic &
  0.293 &
  \bfseries 1.000 &
  0.471 &
  \bfseries 1.000 &
   &
  0.433 &
  \bfseries 1.000 &
  0.748 &
  \bfseries 1.000 &
   &
  0.656 &
  \bfseries 1.000 &
  0.796 &
  \bfseries 1.000 \\
BERTopic$_{\rm{Aug}}$ &
  0.303 &
  \bfseries 1.000 &
  0.468 &
  \bfseries 1.000 &
   &
  0.422 &
  \bfseries 1.000 &
  0.749 &
  \bfseries 1.000 &
   &
  0.637 &
  \bfseries 1.000 &
  0.795 &
  \bfseries 1.000 \\ \hline\hline
\bfseries GPT-3.5$_{\rm{Par}}$ &
  0.213 &
  \bfseries 1.000 &
  0.384 &
  0.994 &
   &
  0.321 &
  0.968 &
  0.585 &
  0.952 &
   &
  0.636 &
  \bfseries 1.000 &
  0.694 &
  \bfseries 1.000 \\ 
\bfseries GPT-3.5$_{\rm{Seq}}$ &
  0.197 &
  0.984 &
  0.335 &
  0.967 &
   &
  0.334 &
  0.975 &
  0.583 &
  0.954 &
   &
  0.479 &
  \bfseries 1.000 &
  0.689 &
  0.994 \\  \hline
\bfseries GPT-4$_{\rm{Par}}$ &
  0.241 &
  \bfseries 1.000 &
  0.402 &
  \bfseries 1.000 &
  \multicolumn{1}{l}{} &
  0.392 &
  \bfseries 1.000 &
  0.661 &
  0.995 &
  \multicolumn{1}{l}{} &
  0.578 &
  \bfseries 1.000 &
  0.754 &
  \bfseries 1.000 \\ 
\bfseries GPT-4$_{\rm{Seq}}$ &
  0.224 &
  0.983 &
  0.403 &
  0.994 &
  \multicolumn{1}{l}{} &
  0.373 &
  \bfseries 1.000 &
  0.660 &
  0.951 &
  \multicolumn{1}{l}{} &
  0.554 &
  0.931 &
  0.626 &
  0.883 \\
\hline
\end{tabular}%
}
\caption{Document coverage (\textit{DC}) and factuality (\textit{Fa}) results under 5 and 15 topics ($K=5$ and $K=15$). Since baseline models without data augmentation discover topics based only on documents, the factuality values are 1.000.}
\label{tab:Coverage-and-Factuality-for-ACL2024}
\end{table*}
\section{Comparison with Conventional Models}
\label{sec:Comparison with Conventional Models}
%We conduct experiments to investigate how well LLMs perform topic modeling compared to conventional models, examining two key aspects: the quality of topics and the impact of LLM-specific concerns, including shortcuts and hallucinations.
We quantitatively compare LLMs to conventional topic models, assessing topic quality and LLM-specific issues like shortcuts and hallucinations.
%%%%%%%%%%%%%%%%%%%%%%%%%%%%%%%%%%%%%%
\subsection{Dataset}
\label{sec:Dataset}
We employ three tokenized datasets provided by ~\citet{zhang_sccl_2021}: GoogleNewsT~\cite{rakib_googlenewst_2020}, Tweet~\cite{yin_tweet_2016}, and StackOverFlow.\footnote{\url{https://www.kaggle.com/competitions/predict-closed-questions-on-stack-overflow/data?select=train.zip}}
%Following \citet{wu_tsctm_2022}, the datasets are preprocessed as follows: ($\mathrm{i}$) characters are converted to lower case; ($\mathrm{ii}$) words of two letters or less are removed; ($\mathrm{iii}$) words appearing less than five times are filtered out.
%
First, the datasets are preprocessed using the preprocessing pipelines by \citet{wu_tsctm_2022}.
Then, we split each preprocessed dataset into subsets for parallel and sequential prompting, setting the size at 1000\footnote{In preliminary experiments, we checked the performance of LLMs under the setting where the subset size is 250, 500, and 1000. See Appendix \ref{sec:appendix_subset_size}.} and truncating the remaining examples.\footnote{The preprocessing details and the final statistics of the datasets used in this study can be found in Appendix \ref{sec:appendix_preprocessing_and_dataset_statistics}}
%Table \ref{table:datasets_statistics} shows the final statistics of the datasets we use.
%
Note that models other than LLMs take the union of subsets as input.%, and each subset contains different examples for each run.
%%%%%%%%%%%%%%%%%%%%%%%%%%%%%%%%%%%%%%
\subsection{Model}
\label{sec:Baseline Models}
In this experiment, we evaluate GPT-3.5 (gpt-3.5-turbo-0125) and GPT-4 (gpt-4-0125-preview) provided by the OpenAI API as representative LLMs.\footnote{In preliminary experiments, we also tried Llama 2~\cite{llama2}, but found it was not controllable enough for its output to be applied to topic modeling. See Appendix \ref{sec:appendix_llama2}.}
%å
For baseline models, we consider three conventional topic models from different paradigms: LDA,\footnote{\url{https://github.com/BobXWu/TopMost}} a state-of-the-art neural topic model (TSCTM,\footnotemark[5] \citet{wu_tsctm_2022}), and the popular clustering-based model (BERTopic,\footnote{\url{https://maartengr.github.io/BERTopic}} \citet{grootendorst_bertopic_2022}). 
Additionally, we report the results of each model with data augmentation.
All hyperparameters follow default implementations.
\footnote{The details can be found in Appendix \ref{sec:appendix_implementation}.}
%

%%%%%%%%%%%%%%%%%%%%%%%%%%%%%%%%%%%%%%
\subsection{Evaluation}
\label{sec:Evaluation}
We evaluate models under the conditions that the number of topics is 5 or 15, and the number of each topic words is 5.
For evaluation metrics, we use two widely used metrics for topic quality and two new metrics to consider the issues of hallucination and shortcuts, i.e., the possibility of outputting topics not included in documents or reflecting only very limited documents.
We run each model five times, each time with a different order of samples, and report the average score.

\paragraph{Topic Coherence and Diversity}
%We use two popular metrics for topic modeling to evaluate the quality of topics discovered by a model: topic coherence and diversity
Following ~\citet{wu_tsctm_2022}, we calculate \textit{Cv}\footnote{\url{https://github.com/dice-group/Palmetto}} \cite{rober_cv_2015} referencing Wikipedia for topic coherence, and \textit{TU} ~\cite{nan_tu_2019} to evaluate the diversity in the inferred topics.
%Following ~\citet{wu_tsctm_2022}, we calculate a coherence value\footnote{\url{https://github.com/dice-group/Palmetto}} (\textit{Cv},~\citealp{rober_cv_2015}) with Wikipedia for topic coherence, and a topic uniqueness (\textit{TU}, ~\citealp{nan_tu_2019}) to evaluate the diversity in the inferred topics.

\paragraph{Document Coverage}
We are concerned that LLMs take shortcuts, inferring topics that reflect only very limited documents.
Thus we propose a metric \textit{document coverage}, which measures the extent to which discovered topics cover documents.
Document coverage is defined as follows:
\begin{displaymath}
    DC = \frac{\#\, (d_{ref} \text{ that contains at least one } w_{topic})}{\#\, (d_{ref})}
\end{displaymath}
where $d_{ref}$ is a document within the reference document collection, and $w_{topic}$ is the topic word constituting the outputted topics.
A higher \textit{DC} means that discovered topics cover more reference documents.
In this experiment, we use the preprocessed datasets without augmentation as references.

\paragraph{Factuality}
Another potential issue is hallucination, where topics discovered by LLMs may not be included in given documents.
%Therefore, we introduce \textit{factuality}, which measures the degree to which topic words are composed from the vocabulary in the reference documents.
%Factuality is defined as follows:
Therefore, we introduce \textit{factuality} as follows:
\begin{displaymath}
Fa = \frac{\#\, (w_{topic} \text{ present in at least one } d_{ref})}{\#\, (w_{topic})}
\end{displaymath}
A higher \textit{Fa} indicates that more topic words are composed from the vocabulary in the reference documents.
%A higher \textit{Fa} indicates that more topic words are extracted from the vocabulary in the reference documents.
%
%ote that the factuality could be less than one in existing topic modeling with data augmentation due to the word substitution by using out-of-vocabulary words of the documents.

% TO DO

%\section{Results and Discussion\todo{}}
\subsection{Results and Discussion}
%%%%%%%%%%%%%%%%%%%%%%%%%%%%%%%%%%%%%%%%%%%%%%%%%%%%%%%%%%%%%%%%%%%%%%%%%%%%
\paragraph{Topic Quality}
Table \ref{tab:Coherence-and-Diversity-for-ACL2024} shows that the topics generated by LLMs are relatively high-quality both in terms of coherence (\textit{Cv}) and diversity (\textit{TU}).\footnote{We show examples of topics in Appendix \ref{sec:appendix_examples_of_topics}.}
In particular, for coherence, GPT-4 achieved the state-of-the-art performance in all settings, with up to 40 \% improvement.
For instance, on GoogleNewsT, the scores have risen by 41\% (from 0.439 to 0.618) and 28\% (from 0.437 to 0.561), respectively, for each setting of the number of topics.

%%%%%%%%%%%%%%%%%%%%%%%%%%%%%%%%%%%%%%%%%%%%%%%%%%%%%%%%%%%%%%%%%%%%%%%%%%%%
%\input{components/table_before_merging}
\paragraph{Document Coverage}
Table \ref{tab:Coverage-and-Factuality-for-ACL2024} reports that LLMs show relatively lower scores for document coverage (\textit{DC}) than the best baseline models.
This means that the topics generated by LLMs often cover fewer documents than those discovered by the conventional models, which suggests the presence of the shortcuts issue.
Note that in both GPT-3.5 and GPT-4, parallel prompting results in higher \textit{DC} scores compared to sequential prompting, implying it might help mitigate the shortcuts issue.\footnote{Further analysis is provided in Appendix \ref{sec:Analysis of Parallel and Sequential Prompting}.}

\paragraph{Factuality}
In Table \ref{tab:Coverage-and-Factuality-for-ACL2024}, LLMs show lower scores for factuality (\textit{Fa}) than the baseline models, particularly ones without augmentation.
This indicates that some topic words generated by LLMs are not included in the documents. %, suggesting the possible risk of hallucinations.
However, their factuality loss is less than 5\% in almost all settings.
%Furthermore, we analyze these non-existent words and find that most of them are not too problematic to mislead topic interpretation; they are synonyms, derivatives, and related words of words in the documents.\footnote{We show examples of not-existent words and analysis details in Appendix \ref{sec:appendix_analysis_on_factuality}.}
Furthermore, we analyze these non-existent words and find that they are mostly synonyms, derivatives, and related words of words in the documents.\footnote{We show examples of not-existent words and analysis details in Appendix \ref{sec:appendix_analysis_on_factuality}.}
This suggests that LLMs might not frequently generate hallucinated topics that could lead to misinterpretation of the content.

%%%%%%%%%%%%%%%%%%%%%%%%%%%%%%%%%%%%%%%%%%%%%%%%%%%%%%%%%%%%%%%%%%%%%%%%%%%%
\begin{comment}
\paragraph{Parallel and Sequential Prompting}
%Comparison between Two Approaches}
Table \ref{tab:Coherence-and-Diversity-for-ACL2024} and Table \ref{tab:Coverage-and-Factuality-for-ACL2024} show that the parallel prompting approach can identify topics with better coherence and document coverage than the sequential prompting one.
%
To analyze the superior performance of the parallel approach, we calculate \textit{Cv} and \textit{DC} of topics before merging.
%
Table \ref{tab:before_merging} shows that \textit{Cv} and \textit{DC} scores before merging degrade from those of the parallel approach, demonstrating that the merging process can improve both their coherence and document coverage.
%
On the other hand, we analyze the transition of topics during sequential approach and then observe that it tends to update the previously identified topic very little due to strict adherence to our instruction, leading to lower document coverage compared to the parallel approach.\footnote{We show examples of their processing and further analysis in Appendix \ref{sec:appendix_processing_details}.}
\end{comment}

\section{Controlled Topic Modeling}
\label{sec:Controlled Topic Modeling}
In this section, we investigate whether specifying the category of topics via prompts can effectively control the category of the generated topics.
\footnote{Note that we also examine the controllability of the number of topics, demonstrating that LLMs can flexibly generate as many topics as specified in prompts. See Appendix \ref{sec:Controllability of the Number of Topics}.}

\paragraph{Settings}
We employ the widely-used 20 NewsGroups\footnote{\url{http://qwone.com/~jason/20Newsgroups/}} dataset annotated with six major topic categories. 
We randomly extract 500 samples and apply the same preprocessing as in Section \ref{sec:Dataset}.
%
%\paragraph{Models}
We evaluate topic modeling with GPT-4 (gpt-4-turbo-2024-04-09) under three different prompt settings: \textbf{Ctrl$_\texttt{[CAT]}$}, \textbf{Base}, and \textbf{Orcl}, as shown in Table \ref{table:prompts}.
Using \textbf{Ctrl$_\texttt{[CAT]}$} prompt, GPT-4 conducts controlled topic modeling focused on the category specified in \texttt{[CAT]}: \texttt{computer}, \texttt{politics}, or \texttt{science}.
Under \textbf{Base} prompt, GPT-4 executes regular topic modeling without any category specification.
Furthermore, as a hypothetical upper bound on the performance with \textbf{Ctrl$_\texttt{[CAT]}$} prompt, we introduce an oracle model with \textbf{Orcl} prompt, which performs topic modeling provided with only documents annotated with \texttt{[CAT]}.

\paragraph{Results and Discussion}
To assess the controllability of GPT-4 with \textbf{Ctrl$_\texttt{[CAT]}$} prompt, we calculate document coverage (\textit{DC}; Section \ref{sec:Evaluation}), using documents in the \texttt{[CAT]} category as reference documents.
Table \ref{tab:controllability_Coverage} demonstrates that GPT-4 achieves higher \textit{DC}$_\texttt{[CAT]}$ scores with \textbf{Ctrl$_\texttt{[CAT]}$} prompt compared to \textbf{Base} prompt, but scores lower than GPT-4 with \textbf{Orcl} prompt.
This suggests that LLMs may have limited controllability in identifying topics related to a specified category in prompts, indicating opportunities for improvement.
Furthermore, to explore the influence of controlled topic modeling via prompts on hallucination issues, we calculate factuality (\textit{Fa}; Section \ref{sec:Evaluation}) using all provided samples as the reference.
Table \ref{tab:controllability_Factuality} shows that \textit{Fa} of GPT-4 when using \textbf{Ctrl$_\texttt{[CAT]}$} prompt is lower than that when using \textbf{Base} prompt, implying that specifying a category in prompts may increase the hallucination risk.

\begin{table}[t!]
\small
\centering
\begin{minipage}[b]{.65\columnwidth}
\centering
\resizebox{\linewidth}{!}{%
\begin{tabular}{lccc}
\toprule%\hline
        Prompt & $DC_{\texttt{com}}$   & $DC_{\texttt{pol}}$   & $DC_{\texttt{sci}}$   \\ \midrule%\hline
Base & 0.621 & 0.530 & 0.651 \\
Orcl  & \textbf{0.968} & \textbf{0.823} & \textbf{0.884} \\ %\hline
\textbf{Ctrl$_\texttt{[CAT]}$} & 0.925 & 0.719 & 0.767 \\ \bottomrule%\hline
\end{tabular}%
}
\subcaption{Document coverage}
\label{tab:controllability_Coverage}
\end{minipage}%
\hspace{.05\columnwidth}%
\begin{minipage}[b]{.3\columnwidth}
\centering
\resizebox{\linewidth}{!}{%
\begin{tabular}{lc}
\toprule%\hline
  Prompt & $Fa$                        \\ \midrule%\hline
Base  & \textbf{0.968}                     \\ %\hline
\textbf{Ctrl$_{\texttt{com}}$} & 0.959                     \\
\textbf{Ctrl$_{\texttt{pol}}$} & \multicolumn{1}{l}{0.936} \\
\textbf{Ctrl$_{\texttt{sci}}$} & 0.895                     \\ \bottomrule%\hline
\end{tabular}%
}
\subcaption{Factuality}
\label{tab:controllability_Factuality}
\end{minipage}
\caption{(a) Document coverage measured for each category and (b) factuality. We show the average of five runs where the number of topics and topic words are set as five.}
%The value of ${DC_\text{\texttt{category}}}$ for Ctrl is the value when the Ctrl prompt has \texttt{category} in \texttt{[Seed]}.
\label{tab:main-table}
\end{table}

\begin{comment}
\begin{table}[t!]
\small
\centering
\begin{minipage}{.65\columnwidth}
\centering
\resizebox{\linewidth}{!}{%
\begin{tabular}{lccc}
\hline
        & \multicolumn{3}{c}{DC} \\
        & com    & pol   & sci   \\ \hline
Default & 0.621  & 0.530 & 0.651 \\
Oracle  & 0.968  & 0.823 & 0.884 \\ \hline
Control & 0.925  & 0.719 & 0.767 \\ \hline
\end{tabular}%
}
\subcaption{Coverage}
\label{tab:my-table1}
\end{minipage}%
\hspace{.05\columnwidth}%
\begin{minipage}{.3\columnwidth}
\centering
\resizebox{\linewidth}{!}{%
\begin{tabular}{clc}
\hline
\multicolumn{1}{l}{}     &     & Fa                        \\ \hline
\multicolumn{1}{l}{Base} &     & 0.968                     \\ \hline
\multirow{3}{*}{Ctrl}    & com & 0.959                     \\
                         & pol & \multicolumn{1}{l}{0.936} \\
                         & sci & 0.895                     \\ \hline
\end{tabular}%
}
\subcaption{Factuality}
\label{tab:my-table2}
\end{minipage}
\caption{Main Caption for Both Tables}
\label{tab:main-table}
\end{table}
\end{comment}

% TO DO

\section{Conclusion}
\label{sec:conclusion}
This paper comprehensively evaluated LLMs for topic modeling, focusing on topic quality, concerns with LLMs, and their controllability in guiding generated topics via prompts.
Our results show that LLMs can identify coherent topics without compromising diversity or facing significant hallucination issues, while they might take shortcuts by focusing only on parts of given documents.
%Furthermore, our study reveals that the controllability of LLMs based on prompts is limited, suggesting room for further improvement.
%
Furthermore, our study provides quantitative evidence indicating that the controllability of LLMs through prompts is limited.

%\section*{Limitation} %ACL-SRWではAppendixに移動
\section{Limitations}
Since we evaluated LLMs using GPT-3.5 and GPT-4, both of which are closed-source models, we could not thoroughly consider whether their pre-training and instruction-tuning datasets might contain the datasets used in this study.
Current open-source LLMs, such as Llama 2, do not sufficiently adhere to the output format needed for evaluation, as shown in Appendix \ref{sec:appendix_llama2}.
We hope future work will evaluate more advanced open-source LLMs through our frameworks and offer more transparent evaluations.
%
%We do not consider it matters whether GPT-3.5 and GPT-4 have seen these datasets in the pre-training; since topic modeling is an unsupervised task, they can not utilize their knowledge about these datasets in our experiment.
%Our approaches perform promisingly in a practical common situation where the number of topics is about 10 but are less reliable with a larger number of topics, such as 50 or 100, due to the difficulty of controlling the LLM's output format. %

\bibliography{anthology,custom}

\begin{thebibliography}{19}
\expandafter\ifx\csname natexlab\endcsname\relax\def\natexlab#1{#1}\fi

\bibitem[{Blei et~al.(2003)Blei, Ng, and Jordan}]{blei_lda_2003}
David~M Blei, Andrew~Y Ng, and Michael~I Jordan. 2003.
\newblock \href {https://www.jmlr.org/papers/volume3/blei03a/blei03a.pdf?ref=https://githubhelp.com} {Latent dirichlet allocation}.
\newblock \emph{Journal of machine Learning research}, 3(Jan):993--1022.

\bibitem[{Churchill and Singh(2022)}]{churchill_topicmodel_evolution_2022}
Rob Churchill and Lisa Singh. 2022.
\newblock \href {https://doi.org/10.1145/3507900} {The evolution of topic modeling}.
\newblock \emph{ACM Comput. Surv.}, 54(10s).

\bibitem[{Devlin et~al.(2019)Devlin, Chang, Lee, and Toutanova}]{devlin_bert_2019}
Jacob Devlin, Ming-Wei Chang, Kenton Lee, and Kristina Toutanova. 2019.
\newblock \href {https://doi.org/10.18653/v1/N19-1423} {{BERT}: Pre-training of deep bidirectional transformers for language understanding}.
\newblock In \emph{Proceedings of the 2019 Conference of the North {A}merican Chapter of the Association for Computational Linguistics: Human Language Technologies, Volume 1 (Long and Short Papers)}, pages 4171--4186, Minneapolis, Minnesota. Association for Computational Linguistics.

\bibitem[{Dieng et~al.(2020)Dieng, Ruiz, and Blei}]{dieng_etm_2020}
Adji~B. Dieng, Francisco J.~R. Ruiz, and David~M. Blei. 2020.
\newblock \href {https://doi.org/10.1162/tacl_a_00325} {{Topic Modeling in Embedding Spaces}}.
\newblock \emph{Transactions of the Association for Computational Linguistics}, 8:439--453.

\bibitem[{Grootendorst(2022)}]{grootendorst_bertopic_2022}
Maarten Grootendorst. 2022.
\newblock \href {https://arxiv.org/abs/2203.05794} {Bertopic: Neural topic modeling with a class-based tf-idf procedure}.
\newblock \emph{Computing Research Repository}, arXiv:2203.05794.
\newblock Version1.

\bibitem[{Kobayashi(2018)}]{kobayashi_contextual_2018}
Sosuke Kobayashi. 2018.
\newblock \href {https://doi.org/10.18653/v1/N18-2072} {Contextual augmentation: Data augmentation by words with paradigmatic relations}.
\newblock In \emph{Proceedings of the 2018 Conference of the North {A}merican Chapter of the Association for Computational Linguistics: Human Language Technologies, Volume 2 (Short Papers)}, pages 452--457, New Orleans, Louisiana. Association for Computational Linguistics.

\bibitem[{Mu et~al.(2024)Mu, Dong, Bontcheva, and Song}]{mu_lrec_2024}
Yida Mu, Chun Dong, Kalina Bontcheva, and Xingyi Song. 2024.
\newblock \href {https://aclanthology.org/2024.lrec-main.887} {Large language models offer an alternative to the traditional approach of topic modelling}.
\newblock In \emph{Proceedings of the 2024 Joint International Conference on Computational Linguistics, Language Resources and Evaluation (LREC-COLING 2024)}, pages 10160--10171, Torino, Italia. ELRA and ICCL.

\bibitem[{Nan et~al.(2019)Nan, Ding, Nallapati, and Xiang}]{nan_tu_2019}
Feng Nan, Ran Ding, Ramesh Nallapati, and Bing Xiang. 2019.
\newblock \href {https://doi.org/10.18653/v1/P19-1640} {Topic modeling with {W}asserstein autoencoders}.
\newblock In \emph{Proceedings of the 57th Annual Meeting of the Association for Computational Linguistics}, pages 6345--6381, Florence, Italy. Association for Computational Linguistics.

\bibitem[{OpenAI(2023)}]{openai_gpt4_2023}
OpenAI. 2023.
\newblock \href {https://arxiv.org/abs/2303.08774} {Gpt-4 technical report}.
\newblock \emph{Computing Research Repository}, arXiv:2303.08774.
\newblock Version 3.

\bibitem[{Ouyang et~al.(2022)Ouyang, Wu, Jiang, Almeida, Wainwright, Mishkin, Zhang, Agarwal, Slama, Ray, Schulman, Hilton, Kelton, Miller, Simens, Askell, Welinder, Christiano, Leike, and Lowe}]{openai_instructgpt_2022}
Long Ouyang, Jeffrey Wu, Xu~Jiang, Diogo Almeida, Carroll Wainwright, Pamela Mishkin, Chong Zhang, Sandhini Agarwal, Katarina Slama, Alex Ray, John Schulman, Jacob Hilton, Fraser Kelton, Luke Miller, Maddie Simens, Amanda Askell, Peter Welinder, Paul~F Christiano, Jan Leike, and Ryan Lowe. 2022.
\newblock \href {https://proceedings.neurips.cc/paper_files/paper/2022/file/b1efde53be364a73914f58805a001731-Paper-Conference.pdf} {Training language models to follow instructions with human feedback}.
\newblock In \emph{Advances in Neural Information Processing Systems}, volume~35, pages 27730--27744.

\bibitem[{Pham et~al.(2023)Pham, Hoyle, Sun, and Iyyer}]{pham_topicgpt_2023}
Chau~Minh Pham, Alexander Hoyle, Simeng Sun, and Mohit Iyyer. 2023.
\newblock \href {http://arxiv.org/abs/2311.01449} {Topicgpt: A prompt-based topic modeling framework}.

\bibitem[{Rakib et~al.(2020)Rakib, Zeh, Jankowska, and Milios}]{rakib_googlenewst_2020}
Md~Rashadul~Hasan Rakib, Norbert Zeh, Magdalena Jankowska, and Evangelos Milios. 2020.
\newblock \href {https://doi.org/https://doi.org/10.1007/978-3-030-51310-8_10} {Enhancement of short text clustering by iterative classification}.
\newblock In \emph{Natural Language Processing and Information Systems}, pages 105--117, Cham. Springer International Publishing.

\bibitem[{R\"{o}der et~al.(2015)R\"{o}der, Both, and Hinneburg}]{rober_cv_2015}
Michael R\"{o}der, Andreas Both, and Alexander Hinneburg. 2015.
\newblock \href {https://doi.org/10.1145/2684822.2685324} {Exploring the space of topic coherence measures}.
\newblock In \emph{Proceedings of the Eighth ACM International Conference on Web Search and Data Mining}, WSDM '15, page 399–408, New York, NY, USA. Association for Computing Machinery.

\bibitem[{Srivastava and Sutton(2017)}]{srivastava_prodlda_2017}
Akash Srivastava and Charles Sutton. 2017.
\newblock \href {https://openreview.net/forum?id=BybtVK9lg} {Autoencoding variational inference for topic models}.
\newblock In \emph{International Conference on Learning Representations}.

\bibitem[{Touvron et~al.(2023)Touvron, Martin, Stone, Albert, Almahairi, Babaei, Bashlykov, Batra, Bhargava, Bhosale, Bikel, Blecher, Ferrer, Chen, Cucurull, Esiobu, Fernandes, Fu, Fu, Fuller, Gao, Goswami, Goyal, Hartshorn, Hosseini, Hou, Inan, Kardas, Kerkez, Khabsa, Kloumann, Korenev, Koura, Lachaux, Lavril, Lee, Liskovich, Lu, Mao, Martinet, Mihaylov, Mishra, Molybog, Nie, Poulton, Reizenstein, Rungta, Saladi, Schelten, Silva, Smith, Subramanian, Tan, Tang, Taylor, Williams, Kuan, Xu, Yan, Zarov, Zhang, Fan, Kambadur, Narang, Rodriguez, Stojnic, Edunov, and Scialom}]{llama2}
Hugo Touvron, Louis Martin, Kevin Stone, Peter Albert, Amjad Almahairi, Yasmine Babaei, Nikolay Bashlykov, Soumya Batra, Prajjwal Bhargava, Shruti Bhosale, Dan Bikel, Lukas Blecher, Cristian~Canton Ferrer, Moya Chen, Guillem Cucurull, David Esiobu, Jude Fernandes, Jeremy Fu, Wenyin Fu, Brian Fuller, Cynthia Gao, Vedanuj Goswami, Naman Goyal, Anthony Hartshorn, Saghar Hosseini, Rui Hou, Hakan Inan, Marcin Kardas, Viktor Kerkez, Madian Khabsa, Isabel Kloumann, Artem Korenev, Punit~Singh Koura, Marie-Anne Lachaux, Thibaut Lavril, Jenya Lee, Diana Liskovich, Yinghai Lu, Yuning Mao, Xavier Martinet, Todor Mihaylov, Pushkar Mishra, Igor Molybog, Yixin Nie, Andrew Poulton, Jeremy Reizenstein, Rashi Rungta, Kalyan Saladi, Alan Schelten, Ruan Silva, Eric~Michael Smith, Ranjan Subramanian, Xiaoqing~Ellen Tan, Binh Tang, Ross Taylor, Adina Williams, Jian~Xiang Kuan, Puxin Xu, Zheng Yan, Iliyan Zarov, Yuchen Zhang, Angela Fan, Melanie Kambadur, Sharan Narang, Aurelien Rodriguez, Robert Stojnic, Sergey Edunov, and Thomas
  Scialom. 2023.
\newblock \href {https://arxiv.org/abs/2307.09288} {Llama 2: Open foundation and fine-tuned chat models}.
\newblock \emph{Computing Research Repository}, arXiv:2307.09288.
\newblock Version 2.

\bibitem[{Wu et~al.(2022)Wu, Luu, and Dong}]{wu_tsctm_2022}
Xiaobao Wu, Anh~Tuan Luu, and Xinshuai Dong. 2022.
\newblock \href {https://doi.org/10.18653/v1/2022.emnlp-main.176} {Mitigating data sparsity for short text topic modeling by topic-semantic contrastive learning}.
\newblock In \emph{Proceedings of the 2022 Conference on Empirical Methods in Natural Language Processing}, pages 2748--2760, Abu Dhabi, United Arab Emirates. Association for Computational Linguistics.

\bibitem[{Yin and Wang(2016)}]{yin_tweet_2016}
Jianhua Yin and Jianyong Wang. 2016.
\newblock \href {https://doi.org/10.1109/ICDE.2016.7498276} {A model-based approach for text clustering with outlier detection}.
\newblock In \emph{2016 IEEE 32nd International Conference on Data Engineering (ICDE)}, pages 625--636.

\bibitem[{Zhang et~al.(2021)Zhang, Nan, Wei, Li, Zhu, McKeown, Nallapati, Arnold, and Xiang}]{zhang_sccl_2021}
Dejiao Zhang, Feng Nan, Xiaokai Wei, Shang-Wen Li, Henghui Zhu, Kathleen McKeown, Ramesh Nallapati, Andrew~O. Arnold, and Bing Xiang. 2021.
\newblock \href {https://doi.org/10.18653/v1/2021.naacl-main.427} {Supporting clustering with contrastive learning}.
\newblock In \emph{Proceedings of the 2021 Conference of the North American Chapter of the Association for Computational Linguistics: Human Language Technologies}, pages 5419--5430, Online. Association for Computational Linguistics.

\bibitem[{Zhang et~al.(2022)Zhang, Fang, Chen, and Namazi~Rad}]{zhang_neural_clustering_2022}
Zihan Zhang, Meng Fang, Ling Chen, and Mohammad~Reza Namazi~Rad. 2022.
\newblock \href {https://doi.org/10.18653/v1/2022.naacl-main.285} {Is neural topic modelling better than clustering? an empirical study on clustering with contextual embeddings for topics}.
\newblock In \emph{Proceedings of the 2022 Conference of the North American Chapter of the Association for Computational Linguistics: Human Language Technologies}, pages 3886--3893, Seattle, United States. Association for Computational Linguistics.

\end{thebibliography}

\appendix
\clearpage

\section{Preliminary Experiments}
\label{sec:Preliminary experiments}
In preliminary experiments, we search prompts and a subset size for our prompting methods to maximize the performance of LLMs.

%%%%%%%%%%%%%%%%%%%%%%%%%%%%%%%%%%%%%%%%%%%%%%%%%%%%%%%%%%%%%%%%%%%%%%%%%%%%
\subsection{Prompts}
\label{sec:appendix_prompts}
We first considered the \textbf{Par$_\text{TM}$} prompt and then proceeded to the \textbf{Par$_\text{Mrg}$} prompt and the \textbf{Seq$_\text{TM}$} prompt.

%%%%%%%%%%%%%%%%%%%%%%%%%%%%%%%%%%%%%%%%%%%%%%%%%%%%%%%%%%%%%%%%%%%%%%%%%%%%
\paragraph{Par$_\text{TM}$}
We checked three kinds of prompts, as shown in Table \ref{table:candidate_for_base_prompts}.
We finally selected a \textbf{Direct} prompt as a tentative a \textbf{Par$_\text{TM}$} prompt, which achieved the highest performance
We then also consider the effect of the insertion of the following phrases, which are expected to improve scores of topic coherence, diversity, and document coverage, respectively. 
\begin{description}
    \item[\textit{Cv}] ``NOTE: Make top words for each topic likely to occur together in the documents''
    \item[\textit{TU}] ``NOTE: Make the top words unique across topics.''
    \item[\textit{DC}] ``NOTE: Maximize the number of documents that contain at least one of the top words.''
\end{description}
However, we found that none of the above can positively influence LLMs’ performance in our methods. 
Therefore, we selected a \textbf{Direct} prompt without phrase insertion as the \textbf{Par$_\text{TM}$} prompt.

%%%%%%%%%%%%%%%%%%%%%%%%%%%%%%%%%%%%%%%%%%%%%%%%%%%%%%%%%%%%%%%%%%%%%%%%%%%%
\paragraph{Par$_\text{Mrg}$}
Considering the Par$_\text{TM}$ prompt, we created a \textbf{Base Par$_\text{Mrg}$} prompt, which has a similar structure to the Par$_\text{TM}$ (Table \ref{table:simplest_prompt_for_parallel_and_sequential}).
Then, we consider the insertion of the following phrases:
\begin{description}
    \item[\textit{Goal}] ``We aim to identify topics for the entire document set by merging the topic modeling results for each subset.''
    \item[\textit{Detail}] ``NOTE: Outputs should reflect the topics before merging as much as possible. Output should contain topics that often appear before merging and not have ones that don't appear much before merging.''
\end{description}
Experimental results showed our methods performed the best when we inserted both the \textit{Goal} phrase and the \textit{Detail} phrase into the \textbf{Base Par$_\text{TM}$}.
 
As a result, we employ a \textbf{Base Par$_\text{TM}$} prompt with both phrases as the \textbf{Par$_\text{TM}$} prompt for the parallel approach.

%%%%%%%%%%%%%%%%%%%%%%%%%%%%%%%%%%%%%%%%%%%%%%%%%%%%%%%%%%%%%%%%%%%%%%%%%%%%
\paragraph{Seq$_\text{TM}$}
Similar to the prompt for parallel, we first created a simple \textbf{Base Seq$_\text{TM}$} prompt for the sequential approach in Table \ref{table:simplest_prompt_for_parallel_and_sequential}, and validated the effect of the insertion of the following phrases.
\begin{description}
    \item[\textit{Goal}] ``We aim to identify topics for the entire document set by sequentially updating tentative topics identified from each subset, considering topics identified just before from another subset.''
    \item[\textit{Detail}] ``NOTE: Outputs should be the same as the previous topics as much as possible. You can change them minimally only when the given documents don't include them much, and a new topic needs to be added to describe the documents.''
\end{description}
We also found that the insertion of both above phrases is most effective in improving the performance of the sequential method.
Thus, we utilize a \textbf{Base Seq$_\text{TM}$} prompt, incorporating both phrases as the \textbf{Seq$_\text{TM}$} prompt for the sequential approach.

%%%%%%%%%%%%%%%%%%%%%%%%%%%%%%%%%%%%%%%%%%%%%%%%%%%%%%%%%%%%%%%%%%%%%%%%%%%%
\subsection{Subset Size}
\label{sec:appendix_subset_size}
We assume candidates of the subset size are 250, 500, and 1000.
It would be difficult to adopt more than 1000 as the subset size due to the context length of GPT-3.5 (gpt-3.5-turbo-0125), which we plan to use for the main experiments.

We run the parallel and the sequential methods with GPT-3.5 on GoogleNewsT in each case of the subset size.
Table \ref{tab:search_for_subset_size} reports the average scores of each method for five runs.
We found the tendency of improving topic coherence as the subset size increase, but could not find any tendency for the other metrics.
We finally select 1000 as the subset size because the performance of each models are relatively high in all metrics under that setting.

Note that our proposed methods with the subset size 250 or 500 could discover competitive or higher-quality topics compared to existing models shown in Table \ref{tab:Coherence-and-Diversity-for-ACL2024} and Table \ref{tab:Coverage-and-Factuality-for-ACL2024}.
This suggests our methods could perform well regardless of the context length of LLMs applied to our methods.

%%%%%%%%%%%%%%%%%%%%%%%%%%%%%%%%%%%%%%%%%%%%%%%%%%%%%%%%%%%%%%%%%%%%%%%%%%%%
\subsection{Llama 2}
\label{sec:appendix_llama2}
In preliminary experiments, we also tried llama-2-7b-chat and llama-2-13b-chat as LLMs for our methods and found that it is difficult for Llama 2 \cite{llama2} to perform topic modeling regardless of prompts and the subset size we use.
Table \ref{tab:llama2_outputs} shows outputs of Llama 2 when given the \textbf{Par$_\text{TM}$} prompt with the subset size 100 on GoogleNewsT.
Llama 2 can not make adequate output following our instruction for the number of topics and topic words, while GPT-3.5 and GPT-4 can perform it steadily under the same setting.

\section{Experimental Details}
%%%%%%%%%%%%%%%%%%%%%%%%%%%%%%%%%%%%%%%%%%%%%%%%%%%%%%%%%%%%%%%%%%%%%%%%%%%%
\subsection{Preprocessing and Dataset Statistics}
\label{sec:appendix_preprocessing_and_dataset_statistics}
Following \citet{wu_tsctm_2022}, the datasets are preprocessed as follows: ($\mathrm{i}$) characters are converted to lower case; ($\mathrm{ii}$) words of two letters or less are removed; ($\mathrm{iii}$) words appearing less than five times are filtered out.
Table \ref{table:datasets_statistics} shows the final statistics of the datasets we use in Section \ref{sec:Comparison with Conventional Models} and Section \ref{sec:Controlled Topic Modeling}.

\subsection{Implementation Details}
\label{sec:appendix_implementation}
We run TSCTM for 200 epochs.
 In the case without data augmentation, we run it with temperature as 0.5 and weight contrast as 1.0. 
In the case of data augmentation, we run it with temperature as 0.07, weight contrast as 3.0, and same quant as 0.001. 
For data augmentation, we applied WordNet\footnote{\url{https://github.com/makcedward/nlpaug}} and Contextual Augmenter\footnotemark[3]~\cite{kobayashi_contextual_2018} with 30\% word replacement, and filtered low-frequency words as in the preprocessing.
Each Augmenter randomly replaces words in an input text with synonyms defined by WordNet and with words predicted by BERT~\cite{devlin_bert_2019}\footnote{\url{https://huggingface.co/bert-base-uncased}}, respectively.
For gpt-3.5-turbo-0125, gpt-4-0125-preview, and BERTopic, we utilized their original configurations without modification.

%%%%%%%%%%%%%%%%%%%%%%%%%%%%%%%%%%%%%%%%%%%%%%%%%%%%%%%%%%%%%%%%%%%%%%%%%%%%
\subsection{Examples of Prompts}
\label{sec:appendix_examples_of_prompts}
Table \ref{table:prompts_examples} shows examples of prompts used in the experiments in Section \ref{sec:Comparison with Conventional Models}.

\section{Result Details}
%%%%%%%%%%%%%%%%%%%%%%%%%%%%%%%%%%%%%%%%%%%%%%%%%%%%%%%%%%%%%%%%%%%%%%%%%%%%
\subsection{Examples of Topics}
\label{sec:appendix_examples_of_topics}
Following \citet{wu_tsctm_2022}, we randomly select some examples of topics identified by LDA, BERTopic, and our proposed methods with GPT-4.

%%%%%%%%%%%%%%%%%%%%%%%%%%%%%%%%%%%%%%%%%%%%%%%%%%%%%%%%%%%%%%%%%%%%%%%%%%%%
\subsection{Analysis of Parallel and Sequential Prompting}
\label{sec:Analysis of Parallel and Sequential Prompting}
Table \ref{tab:Coherence-and-Diversity-for-ACL2024} and Table \ref{tab:Coverage-and-Factuality-for-ACL2024} show that the parallel prompting approach can identify topics with better coherence and document coverage than the sequential prompting one.
To analyze the superior performance of the parallel approach, we calculate \textit{Cv} and \textit{DC} of topics before merging.
Table \ref{tab:before_merging} shows that \textit{Cv} and \textit{DC} scores before merging degrade from those of the parallel approach, demonstrating that the merging process can improve both their coherence and document coverage.

Additionally, we examine the transition of topics during their processing and then observe that the sequential approach tends to update the previously identified topic very little due to strict adherence to our instruction, leading to lower document coverage compared to the parallel approach.
Table \ref{tab:topic_for_each_subset_and_output} reports the concrete examples of topics identified for each subset and the final output to demonstrate the processing in our methods.
In the parallel approach, we find that LLM reasonably merges topics from each subset.
For instance, bold topics in each subset are merged into one topic in the final output, using words from both subsets.
In contrast, in the sequential approach, the final output is the same as the topics for the first subset except for only one pair of bold words.
This indicates that LLMs with the the sequential approach could too strictly keep topics for the previous subset so that they can not output topics that sufficiently reflect the entire set.

%%%%%%%%%%%%%%%%%%%%%%%%%%%%%%%%%%%%%%%%%%%%%%%%%%%%%%%%%%%%%%%%%%%%%%%%%%%%
\subsection{Examples of Topic Words Not Included in the Documents}
\label{sec:appendix_analysis_on_factuality}
Table \ref{table:hallucinations} shows examples of words not included in the documents outputted in topic modeling on GoogleNewsT.
GPT-3.5 output in bold are the names of entities (e.g., \textbf{broncos}, \textbf{gree}, and \textbf{watson}) or words that do not exist in the real world (e.g., \textbf{dorffiefskee}).
Such words are considered harmful because they may induce misinterpretation of topics. 
However, only a small number of such words were found, and most of them were synonyms, synonyms, derivatives, or related words in the documents.

\section{Controllability of the Number of Topics}
\label{sec:Controllability of the Number of Topics}
Unlike conventional topic models that consistently produce the specified number of topics, LLMs may not generate the exact number of topics indicated in the prompt.
Therefore, to evaluate LLM's controllability of the number of topics, we investigate its output when the specified number of topics in prompts is set to 5, 10, 15, and 20, while keeping the number of topic words fixed at 5.
We use the documents set from Section \ref{sec:Controlled Topic Modeling} and GPT-4 (gpt-4-turbo-2024-04-09) with \textbf{Base} prompt, running ten trials for each setting.
The results show that the same number of topics as specified are outputted in all trials, demonstrating GPT-4's sufficient controllability of the number of topics.

%\input{sections/limitation} %ACL-SRWではappendixに移動

%%%%%%%%%%%%%%%%%%%%%%%%%%%%%%%%%%%%%%%%%%%%%%%%%%%%%%%%%%%%%%%%%%%%%%%%%%%%
\begin{table*}[t!]
    \small
    \center
    {%\setlength{tabrowsep}{1pt}
    \begin{tabular}{cl}
    \toprule
     ID & Candidates of Base Prompt Template\\
    \midrule
    \textbf{Direct} &
    \begin{minipage}{0.80\textwidth}
    \textbf{Write the results of simulating topic modeling for the following documents}, each starting with "\#."\\
    Assume you will finally identify \texttt{[NUM\_TOPICS]} topics and use 5 top words for each topic.\\
    NOTE: Outputs must always be in the format "Topic k: word word word word word" and nothing else.\\
    """ \\
    $\texttt{[DOCS]}$\\
    """
    \end{minipage} \\
    
    \midrule 
    
    \textbf{Indirect} &
    \begin{minipage}{0.80\textwidth}
    Discover latent \texttt{[NUM\_TOPICS]} topics in the following documents, each starting with "\#."\\
    For each topic, write 5 words extracted from input texts to show its meanings.\\
    NOTE: Outputs must always be in the format "Topic k: word word word word word" and nothing else.\\
    """\\
    $\texttt{[DOCS]}$\\
    """
    \end{minipage} \\
    
    \midrule
    
    \textbf{Direct$_\text{reverse}$} &
    \begin{minipage}{0.80\textwidth}
    """ \\
    $\texttt{[DOCS]}$\\
    """\\
    \textbf{Write the results of simulating topic modeling for the above documents}, each starting with "\#."\\
    Assume you will finally identify \texttt{[NUM\_TOPICS]} topics and use 5 top words for each topic.\\
    NOTE: Outputs must always be in the format "Topic k: word word word word word" and nothing else.
    \end{minipage} \\
    \bottomrule
    \end{tabular}
    }
    
    \caption{Candidate prompts for Par$_\text{TM}$. \texttt{[DOCS]} and \texttt{[NUM\_TOPICS]} are replaced by a subset of documents and by the nuber of topics.}
    \label{table:candidate_for_base_prompts}
\end{table*}
\begin{table*}[t!]
    \small
    \center
    {%\setlength{tabrowsep}{1pt}
    \begin{tabular}{cl}
    \toprule
     ID & Candidates of Base Prompt Template\\
    \midrule
    \textbf{Base Par$_\text{TM}$} &
    \begin{minipage}{0.80\textwidth}
    Write the results of merging the following topic modeling results for each subset of the document set.\\
    Each result starts with "- n" and its topics start with "\#"\\
    """\\
    - 1\\
    $\texttt{[TOPICS]}$\\
    \\
    - 2\\
    $\texttt{[TOPICS]}$\\
    \\
    - 3\\
    ...\\
    """
    \end{minipage} \\

    \midrule
    
    \textbf{Base Seq$_\text{TM}$} &
    \begin{minipage}{0.80\textwidth}
    Write the results of simulating topic modeling for the following documents, each starting with "\#."\\ 
    Make the most use of the following topics previously identified from another set of documents, each starting with "Topic k:":\\
    """\\
    $\texttt{[TOPICS]}$\\
    """\\
    Assume you will finally identify \texttt{[NUM\_TOPICS]} topics and use 5 top words for each topic.\\
    NOTE: Outputs must always be in the format "Topic k: word word word word word" and nothing else.\\
    """\\
    $\texttt{[DOCS]}$\\
    """
    \end{minipage} \\
    \bottomrule
    \end{tabular}
    }
    
    \caption{Base prompts for the parallel and the sequential. \texttt{[DOCS]}, \texttt{[TOPICS]}, and \texttt{[NUM\_TOPICS]} are replaced by a subset of documents, previously identified topics, and the nuber of topics, respectively.}
    \label{table:simplest_prompt_for_parallel_and_sequential}
\end{table*}
\begin{table*}[h!]
\centering
\begin{minipage}{0.45\textwidth}
\centering
\begin{tabular}{lcccc}
\hline
Subset Size & \textit{Cv}             & \textit{TU}            & \textit{DC}             & \textit{Fa}            \\ \hline
$250$     & 0.531          & 0.936         & \textbf{0.241} & \textbf{1.000} \\
$500$     & \textbf{0.572} & 0.896         & \textbf{0.241} & \textbf{1.000} \\
$1000$    & 0.571          & \textbf{0.960} & 0.213          & \textbf{1.000} \\ \hline
\end{tabular}%
\subcaption{Parallel}
\label{tab:table1}
\end{minipage}%
\hfill
\begin{minipage}{0.45\textwidth}
\centering
\begin{tabular}{lcccc}
\hline
Subset Size & \textit{Cv}    & \textit{TU}    & \textit{DC}    & \textit{Fa}    \\ \hline
$250$     & 0.524 & 0.976 & \textbf{0.198} & \textbf{0.992} \\
$500$     & 0.529 & \textbf{0.992} & 0.193 & 0.976 \\
$1000$    & \textbf{0.562} & 0.984 & 0.197 & 0.984 \\ \hline
\end{tabular}%
\subcaption{Sequential}
\end{minipage}
\caption{Results of the parallel and the sequential under five topics on GoogleNewsT when changing the subset size among 250, 500, and 1000. The best scores are in \textbf{bold}.}
\label{tab:search_for_subset_size}
\end{table*}

\begin{table*}[t!]
    \small
    \center
    {%\setlength{tabrowsep}{1pt}
    \begin{tabular}{cl}
    \toprule
     Model size & Examples of Llama2 Output\\
    \midrule
    
    \textbf{7B} &
    \begin{minipage}{0.80\textwidth}
    Topic 1:\\
    Top words: relief, challenge, face\\
    Topic 2:\\
    Top words: welker, concussion, test\\
    Topic 3:\\
    Top words: live, stream, champion, league\\
    Topic 4:\\
    Top words: bargain, black, friday, shopping\\
    Topic 5:\\
    Top words: scotland, independence, white, paper\\
    Note: Each topic is represented by 5 top words, which are the most frequently occurring words in the given documents.
    \end{minipage} \\

    \midrule
    
    \textbf{13B} &
    \begin{minipage}{0.80\textwidth}
    Topic 1: Disasters and Relief Efforts\\
    Topic 2: Sports and Injuries\\
    Topic 3: Technology and Gadgets\\
    Topic 4: Politics and Leadership\\
    Topic 5: Entertainment and Celebrities
    \end{minipage} \\

    \bottomrule
    
    \end{tabular}
    }
    
    \caption{Examples of Llama 2 outputs when we provide Par$_\text{TM}$ on GoogleNewsT under the conditions that the number of topics and topic words is five and the subset size is 100.}
    \label{tab:llama2_outputs}
\end{table*}
%%%%%%%%%%%%%%%%%%%%%%%%%%%%%%%%%%%%%%%%%%%%%%%%%%%%%%%%%%%%%%%%%%%%%%%%%%%%
\begin{table*}[t!]
    %\small
    \center
    \begin{tabular}{lrrrr}
    \toprule
    
             & \# of & Text  & Vocabulary \\
     Dataset & Documents & Length & Size\\
     
    \midrule
    
    Tweet & \hphantom{0}2000 & 5.47 & \hphantom{0}706 \\ 
    %\midrule
    GoogleNewsT & 11000 & 5.25 & 2376 \\ 
    %\midrule
    StackOverFlow & 19000 & 4.71 & 2544 \\ 
    
    \midrule
    
    20NG (all samples) & \hphantom{00}500 & 147.25 & 2095 \\ 
    %\midrule
    20NG (\texttt{computer}) & \hphantom{00}126 & 98.27 & 1437 \\ 
    %\midrule
    20NG (\texttt{politics}) & \hphantom{000}77 & 296.69 & 1792 \\ 
    %\midrule
    20NG (\texttt{science}) & \hphantom{000}57 & 113.51 & 1672 \\ 
    
    \bottomrule
    \end{tabular}
      %\caption{Statistics of the datasets after preprocessing with and without data augmentation (Aug). $|D|$ means the number of documents, $Len$ means the average number of words per document, and $|V|$ means its vocabulary size. The number of documents before data augmentation is less than 1000 due to the filtering of low-frequency words.}
      \caption{Dataset statistics. Each value is the average for five runs.}
    \label{table:datasets_statistics}
    % \vspace{-0.5cm} 
\end{table*}
\begin{table*}[t!]
    \small
    \center
    {%\setlength{tabrowsep}{1pt}
    \begin{tabular}{cl}
    \toprule
     ID & Prompt Example\\
    \midrule
    \textbf{Par$_\text{TM}$} &
    \begin{minipage}{0.80\textwidth}
    \textbf{Write the results of simulating topic modeling for the following documents}, each starting with "\#."\\
    Assume you will identify 5 topics and use 5 top words for each topic.\\
    NOTE: Outputs must always be in the format "Topic k: word word word word word" and nothing else.\\
    """ \\
    \# philippine typhoon relief effort face challenge\\
    \# wes welker concussion test bronco\\
    \# basel chelsea live stream champion league watch\\
    ... \\
    \# discus black friday shopping secret\\
    """
    \end{minipage} \\
    
    \midrule 
    
    \textbf{Par$_\text{Mrg}$} &
    \begin{minipage}{0.80\textwidth}
    We aim to identify topics for the entire document set by merging the topic modeling results for each subset.\\
    \textbf{Write the results of merging the following topic modeling results for each subset of the document set.}\\
    Each result starts with "- n" and its topics start with "\#"\\
    """\\
    - 1\\
    \# comet ison thanksgiving sun solar\\
    \# kanye west bound parody video\\
    \# nokia lumia release mobile device\\
    \# black friday shopping thanksgiving sale\\
    \# alec baldwin msnbc cancellation defends\\
    \\
    ...\\
    \\
    - 11\\
    \# nokia lumia sale december phone\\
    \# kanye west kim kardashian taylor\\
    \# black friday deal best sales\\
    \# irs rule political activity tax\\
    \# bronco patriot win game rivalry\\
    """\\
    Assume you will finally identify 5 topics and use 5 top words for each topic.\\
    NOTE: Outputs should reflect the topics before merging as much as possible. Output should contain topics that often appear before merging and not have ones that don't appear much before merging.\\
    NOTE: Outputs must always be in the format "Topic k: word word word word word" and nothing else.
    \end{minipage} \\
    
    \midrule
    
    \textbf{Seq$_\text{TM}$} &
    \begin{minipage}{0.80\textwidth}
    We aim to identify topics for the entire document set by sequentially updating tentative topics identified from each subset, considering topics identified just before from another subset.\\
    Write the results of simulating topic modeling for the following documents, each starting with "\#." \\
    \textbf{Make the most use of the following topics previously identified from another set of documents, each starting with "Topic k:":}\\
    """\\
    Topic 1: kanye west kim kardashian bound\\
    Topic 2: xbox black friday cyber monday\\
    Topic 3: hewlett packard nokia lumia company\\
    Topic 4: dancing star finale winner season\\
    Topic 5: syria peace talk china air\\
    """\\
    Assume you will finally identify 5 topics and use 5 top words for each topic.\\
    NOTE: Outputs should be the same as the previous topics as much as possible. You can change them minimally only when the given documents don't include them much, and a new topic needs to be added to describe the documents.\\
    NOTE: Outputs must always be in the format "Topic k: word word word word word" and nothing else.\\
    """\\
    \# spacex falcon launch attempt\\
    \# taylor swift princess gown winter white\\
    \# redbox instant window phone appears nokia exclusive\\
    ...\\
    \# google backed company selling dna analysis kit ordered sale\\
    """
    \end{minipage} \\
    \bottomrule
    \end{tabular}
    }
    
    \caption{Examples of prompts used as Par$_\text{TM}$, Par$_\text{Mrg}$, and Seq$_\text{TM}$ for topic modeling on GoogleNewsT under five topics.}
    \label{table:prompts_examples}
\end{table*}
% \usepackage{multirow}
% \usepackage{graphicx}
\begin{table*}[t]
\centering
%\scalebox{0.96}[0.96]{%
\begin{tabular}{ll}
\toprule
\multicolumn{1}{l}{Model} & Examples of topics                             \\ \hline

\multirow{3}{*}{LDA} 
%    & white men bank paper wearhouse  \\
    & xbox microsoft game patriot bronco    \\
%    & talk scotland independence syria peace  \\ 
    & nokia lumia oldboy launch google \\
    & kobe bryant chelsea lakers basel \\ \hline

\multirow{3}{*}{TSCTM} 
%    & coalition pope berlusconi francis merkel  \\
    & macy parade hanukkah thanksgiving travel   \\
%    & climate hewlett raptor earnings sandy  \\ 
    & china zone african japan johansson \\
    & bronco patriot packer welker illinois \\ \hline
    
\multirow{3}{*}{BERTopic} 
%    & independence pope scotland irs scottish  \\
    & china zone air nsa porn    \\
%    & hiv woman nigella lawson pill  \\ 
    & methane ant emission fire burning \\
    & thanksgiving friday black comet parade \\ \hline
    
\multirow{3}{*}{GPT-4$_{\text{Seq}}$} 
%     & syria peace talk january long\\
     & wes welker nfl concussion game\\
%     & irs rule tax exempt political\\
     & nokia lumia window phone december\\
     & nfl season game player concussion\\ \hline
     
\multirow{3}{*}{GPT-4$_{\text{Par}}$} 
%     & china air defense zone dispute\\
     & san andreas mobile game release\\
%     & nsa porn habit spied radicalizers\\
     & nokia lumia tablet smartphone launch\\
     & thanksgivukkah hanukkah holiday feast rare\\
\bottomrule
\end{tabular}%
%}
\caption{Examples of topics discovered from GoogleNewsT under 15 topics.}
\label{tab:topic_examples}
\end{table*}
% \usepackage{multirow}
% \usepackage{graphicx}
\begin{table*}[]
%\resizebox{\columnwidth}{!}{%
\centering
\begin{tabular}{cclcclcll}
\hline
\multirow{3}{*}{Model} &
  \multicolumn{2}{c}{Tweet} &
   &
  \multicolumn{2}{c}{GoogleNewsT} &
   &
  \multicolumn{2}{l}{StackOverFlow} \\ \cline{2-3} \cline{5-6} \cline{8-9} 
 &
  \multicolumn{2}{c}{$K=15$} &
   &
  \multicolumn{2}{c}{$K=15$} &
   &
  \multicolumn{2}{c}{$K=15$} \\
 &
  \textit{Cv} &
  \multicolumn{1}{c}{\textit{DC}} &
  \textit{\textbf{}} &
  \textit{Cv} &
  \multicolumn{1}{c}{\textit{DC}} &
  \textit{\textbf{}} &
  \multicolumn{1}{c}{\textit{Cv}} &
  \multicolumn{1}{c}{\textit{DC}} \\ \hline
%\multicolumn{1}{l}{GPT-3.5$_{\text{Par w/o Mrg}}$} &
\multicolumn{1}{l}{GPT-3.5} &
  \multicolumn{1}{l}{0.532} &
  0.366 &
  \multicolumn{1}{r}{} &
  \multicolumn{1}{l}{0.517} &
  0.569 &
  \multicolumn{1}{r}{} &
  0.464 &
  0.634 \\
%\multicolumn{1}{l}{GPT-4$_{\text{Par w/o Mrg}}$} &
\multicolumn{1}{l}{GPT-4} &
  \multicolumn{1}{l}{0.580} &
  0.395 &
  \multicolumn{1}{l}{} &
  \multicolumn{1}{l}{0.523} &
  0.665 &
  \multicolumn{1}{l}{} &
  0.519 &
  0.747 \\ \hline
\end{tabular}%
%}
\caption{Coherence (\textit{Cv}) and document coverage (\textit{DC}) of the generated topics in the parallel prompting without the merging process under 15 topics ($K=15$). We take the average of the values computed for each subset in five runs.}
\label{tab:before_merging}
\end{table*}
\begin{table*}[t!]
\centering
\begin{minipage}{0.45\textwidth}
    \centering
    \begin{tabular}{l}
    \hline
    \textbf{Subset 1}\\
    \hline
    fishing fish bass fly report\\
    \textbf{superbowl commercial bowl super best}\\
    king speech oscar nomination award\\
    facebook privacy setting user change\\
    acai berry weight loss diet plan  \\
    \hline
    \\
    \hline
    \textbf{Subset 2}\\
    \hline
    fishing fish fly book saltwater\\
    \textbf{superbowl commercial doritos pepsi volkswagen}\\
    king speech oscar nomination award best\\
    acai berry weight loss diet plan\\
    christina aguilera national anthem super bowl\\
    \hline
    \\
    \hline
    \textbf{Final Output}\\
    \hline
    fishing fish fly bass saltwater\\
    \textbf{superbowl commercial bowl pepsi doritos}\\
    king speech oscar nomination award\\
    acai berry weight loss diet health\\
    facebook privacy setting user change\\
    \hline
    \end{tabular}%
    \subcaption{Parallel}
    \label{tab:table1}
\end{minipage}%
\hfill
\begin{minipage}{0.45\textwidth}
    \centering
    \begin{tabular}{l}
    \hline
    \textbf{Subset 1}\\
    \hline
    fishing commercial superbowl fly bass\\
    facebook privacy setting user \textbf{setting}\\
    king speech oscar nomination award\\
    berry acai weight diet loss\\
    christina aguilera national anthem super\\
    \hline
    \\
    \hline
    \textbf{Final Output}\\
    \hline
    fishing fly superbowl commercial bass\\
    facebook privacy setting user \textbf{security}\\
    king speech oscar nomination award\\
    acai berry weight diet loss\\
    christina aguilera national anthem super\\
    \hline
    \end{tabular}%
    \subcaption{Sequential}
    \label{tab:topic_for_each_subset_and_output}
\end{minipage}
\caption{Topics identified for each subset and the final output by each method with GPT-4 on Tweet under five topics. \textbf{Bold words} are mentioned in Appendix \ref{sec:Analysis of Parallel and Sequential Prompting}}
\label{tab:topic_for_each_subset_and_output}
\end{table*}

\begin{table*}[t]
    %\small
    \centering
    \scalebox{0.96}[0.96]{
    \begin{tabular}{cl}
    
    \toprule
     Model & Examples of the topic words not included in the documents \\
    % & Documents & Length & Vocabulary \\
    \midrule
    TACTM$_{\text{Aug}}$ 
     & twelvemonth sink railway blowout\\ 
     
    %\rule{0pt}{2.5ex}  
    
    \bfseries GPT-3.5$_{\text{Seq}}$
     & \textbf{dorffiefskee} \textbf{broncos} patriots health advancement ocean guilty france legal attorney \\
     
     %\rule{0pt}{2.5ex}  
     
    \textbf{GPT-3.5$_{\text{Par}}$} 
     & \textbf{gree} \textbf{watson} advertisement boat funding attorney declared refugees crash digital\\
     
    \bottomrule
    \end{tabular}
    }
    \caption{Examples of topic words not included in the documents when topic modeling on GoogleNewsT.}
    \label{table:hallucinations}
    % \vspace{-0.5cm} 
\end{table*}

\end{document}